\documentclass[english]{article}
\usepackage[T1]{fontenc}
\usepackage[latin9]{inputenc}
\usepackage{amsmath}
\usepackage{graphicx}
\usepackage{amssymb}

\makeatletter

\providecommand{\tabularnewline}{\\}

\makeatother

\usepackage{babel}
\date{}
\begin{document}

\title{Abstraction Super-structuring Normal Forms: Towards a Theory of Structural
Induction}

\author{Adrian Silvescu and Vasant Honavar}
\maketitle
\begin{center}
\textbf{Technical report}
\end{center}

\begin{center}
Department of Computer Science, Iowa State University, Ames, IA, USA
\end{center}
\begin{abstract}
Induction is the process by which we obtain predictive laws or theories
or models of the world. We consider the structural aspect of induction.
We answer the question as to whether we can find a finite and minmalistic
set of operations on structural elements in terms of which any theory
can be expressed. We identify \textit{abstraction} (grouping \textit{similar}
entities) and super-structuring (combining topologically e.g., spatio-temporally
close entities) as the essential structural operations in the induction
process. We show that only two more structural operations, namely,
\textit{reverse abstraction} and \textit{reverse super-structuring}
(the duals of abstraction and super-structuring respectively) suffice
in order to exploit the full power of Turing-equivalent generative
grammars in induction. We explore the implications of this theorem
with respect to the nature of hidden variables, radical positivism
and the 2-century old claim of David Hume about the principles of
\textit{connexion} among ideas. 
\end{abstract}

\section{Introduction}

The logic of induction, the process by which we obtain predictive
laws, theories, or models of the world, has been a long standing concern
of philosophy, science, statistics and artifical intelligence. Theories
typically have two aspects: structural or qualitative (corresponding
to concepts or variables and their relationships, or, in philosophical
parlance, \textit{ontology}) and numeric or quantitative (corresponding
to parameters e.g., probabilities). Once the qualitative aspect of
a certain law is fixed, the quantitative aspect becomes the subject
of experimental science and statistics. Induction is the process of
inferring predictive laws, theories, or models of the world from a
stream of observations. In general, the observations may be passive,
or may be the outcomes of interventions by the learning agent. Here,
we limit ourselves to induction from passive observation alone. 

Under the \textit{computationalistic assumption} (i.e., the Church-Turing
thesis, which asserts that any theory can be described by a Turing
Machine \cite{Turing 1936}), one way to solve the induction problem
is to enumerate all the Turing machines (dovetailing in order to cope
with the countably infinite number of them) and pick one that strikes
a good balance between the predictability (of the finite experience
stream) and size (complexity) \cite{Solomonoff 1964a}, \cite{Solomonoff 1964b},
\cite{Schmidhuber et al. 1997} or within a Bayesian setting, using
a weighted vote among the predictions of the various models \cite{Hutter 2005}
(See\cite{Burgin 2005} and references therein). In the general setting,
a priori the number of types of possible structural laws that can
be postulated is infinite. This makes it difficult to design general
purpose induction strategy. We ask whether a finite and minimalistic
set of fundamental structural operations suffice to construct \textit{any}
set of laws. If such a set will render induction more tractable because
at any step the learner will have to pick from a small \textit{finite}
set of possible operations as opposed to an infinite one. 

Because Turing machines are rather opaque from a structural standpoint,
we use the alternative, yet equivalent, mechanism of generative grammars%
\footnote{See \cite{Oates et al. 2004} for a similarly motivated attempt using
\textit{Lambda calculus}%
}. This allows us to work with theories that can be built recursively
by applying structural operations drawn from a finite set. The intuition
behind this approach is that induction involves incrementally constructing
complex structures using simpler structures (e.g., using super-structuring,
also called \textit{chunking}), and simplifying complex structures
when possible (e.g., using abstraction). Such a compositional approach
to induction offers the advantage of increased transparency over the
enumerate-and-select approach pioneered by Solomonoff \cite{Solomonoff 1964a},
\cite{Solomonoff 1964b}. It also offers the possibility of reusing
intermediate structures as opposed to starting afresh with a new Turing
machine at each iteration, thereby replacing enumeration by a process
akin to dynamic programming or its heuristic variants such as the
A{*} algorithm. 

We seek laws or patterns that explain a stream of observations through
successive applications of operations drawn from a small finite set.
The induced patterns are not necessarily described solely in terms
of the input observations, but may also use (a finite number of) additional
internal or hidden (i.e., not directly observable) entities. The role
of these internal variables is to simplify explanation. The introduction
of internal variables to aid the explanation process is not without
perils\cite{Popper 1934}%
\footnote{Consider for example, a hidden variable which stands for the truth
value of the sentence: {}``In heaven, if it rains, do the angels
get wet or not?''%
}. One way to preclude the introduction of internal variables is to
apply the following\textit{ demarcation criterion}: If the agent cannot
distinguish possible streams of observations based on the values of
an internal variable, then the variable is non-sensical (i.e., independent
of the data or {}``senses'') %
\footnote{This is a radical interpretation of an idea that shows up in the history
of Philosophy from Positivism through the empiricists and scholastics
down to Aristotle's \emph{{}``Nihil est in intellectu quod non prius
fuerit in sensu''}.%
}. The direct connection requirement restricts the no-nonsense theories
to those formed out empirical laws \cite{Ayer 1936} (i.e, laws that
relate only measurable quantities). However several scientists, including
Albert Einstein, while being sympathetic to the positivists ideas,
have successfully used in their theories, hidden variables that have
at best indirect connection to observables. This has led to a series
of revisions of the positivists doctrine culminating in Carnap's attempt
to accommodate hidden variables in scientific explanations\cite{Carnap 1966}.
The observables and the internal variables in terms of which the explanation
is offered can be seen as the ontology%
\footnote{The ontology in this case is not universal as it is often the case
in philosophy; it is just a set of concepts and interrelations among
them that afford the expression of theories.%
} - i.e., the set of concepts and their interrelationships found useful
by the agent in theorizing about its experience. In this setting,
structural induction is tantamount to ontology construction. 

The rest of the paper is organized as follows: Section 2 introduces
Abstraction Super-structuring Normal Forms that correspond to a general
class of Turing-equivalent generative grammars that can be used to
express theories about the world; and shows that: \textit{abstraction}
(grouping \textit{similar} entities) and super-structuring (combining
topologically e.g., spatio-temporally close entities) as the essential
structural operations in the induction process; Only two more structural
operations, namely, \textit{reverse abstraction} and \textit{reverse
super-structuring} (the duals of abstraction and super-structuring
respectively, suffice in order to exploit the full power of Turing-equivalent
generative grammars in induction. Section 3 interprets the theoretical
results in a larger context the nature of hidden variables, radical
positivism and the 2-century old claim of David Hume about the principles
of \textit{connexion} among ideas. Section 4 concludes with a summary.

\vspace*{-0.1in}

\section{Abstraction Super-Structuring Normal Forms}

We start by recapitulating the definitions and notations for generative
grammars and the theorem that claims the equivalence between Generative
Grammars and Turing Machines. We then draw the connections between
the process of induction and the formalism of generative grammars
and motivate the quest for a minimalistic set of fundamental structural
operations. We then get to the main results of the paper: a series
of characterization theorems of two important classes of Generative
Grammars: Context-Free and General Grammars, in terms of a small set
of fundamental structural operations.

\subsection{Generative Grammars and Turing Machines}

\textbf{Definitions (Grammar)} A (generative) grammar is a quadruple
$(N,T,S,R)$ where $N$ and $T$ are disjoint finite sets called NonTerminals
and Terminals, respectively, $S$ is a distinguished element from
$N$ called the start symbol and $R$ is a set of rewrite rules (a.k.a.
production rules) of the form $(l\to r)$ where $l\in(N\cup T)^{*}N(N\cup T)^{*}$
and $r\in(N\cup T)^{*}$. Additionally, we call $l$ the left hand
side (lhs) and $r$ the right hand side (rhs) of the rule $(l\to r)$.
The language generated by a grammar is defined by $L(G)=\{w\in T^{*}|S\overset{*}{\to}w\}$
where $\overset{*}{\to}$ stands for the reflexive transitive closure
of the rules from $R$. Furthermore $\overset{+}{\to}$ stands for
the transitive (but not reflexive) closure of the rules from $R$.
We say that two grammars $G$,$G'$ are equivalent if $L(G)=L(G')$.
The steps contained in a set of transitions $\alpha\overset{*}{\to}\beta$
is called a derivation. If we want to distinguish between derivations
in different grammars we will write $\alpha\overset{*}{\to_{G}}\beta$
or mention it explicitly. We denote by $\epsilon$ the empty string
in the language. We will sometimes use the shorthand notation $l\to r_{1}|r_{2}|...|r_{n}$\textbf{
}to stand for the set of rules $\{l\to r_{i}\}_{i=1,n}$. See e.g.,
\cite{Salomaa 1985} for more details and examples.

\textbf{Definition (Grammar Types)} Let $G=(N,T,S,R)$ be a grammar.
Then
\begin{enumerate}
\item G is a \textbf{regular grammar (REG)} if all the rules $(l\to r)\in R$
have the property that $l\in N$ and $r\in(T^{*}\cup T^{*}N)$.
\item G is \textbf{context-free grammar (CFG)} if all the rules $(l\to r)\in R$
have the property that $l\in N$.
\item G is \textbf{context-sensitive grammar (CSG)} if all the rules $(l\to r)\in R$
have the property that they are of the form $\alpha A\beta\to\alpha\gamma\beta$
where $A\in N$ and $\alpha,\beta,\gamma\in(N\cup T)^{*}$ and $\gamma\neq\epsilon$.
Furthermore if $\epsilon$ is an element of the language one rule
of the form $S\to\epsilon$ is allowed and furthermore the restriction
that $S$ does not appear in the right hand side of any rule is imposed.
We will call such a sentence an $\epsilon-Amendment$.
\item G is \textbf{general grammar (GG)} if all the rules $(l\to r)\in R$
have no additional restrictions.
\end{enumerate}
\textbf{Theorem 1.} \emph{The set of General Grammars are equivalent
in power with the set of Turing Machines. That is, for every Turing
Machine $T$ there exists a General Grammar $G$ such that $L(G)=L(T)$
and vice versa.}

\emph{Proof.} This theorem is a well known result. See for example
\cite{Salomaa 1985} for a proof%
\footnote{Similar results of equivalence exist for transductive versions of
Turing machines and grammars as opposed to the recognition versions
given here (See e.g., \cite{Burgin 2005} and references therein).
Without loss of generality, we will assume the recognition as opposed
to the transductive setting.%
}. $\Box$

\vspace*{-0.1in}

\subsection{Structural Induction, Generative Grammars and Motivation}

Before proceeding with the main results of the paper we examine the
connections between the setting of generative grammars and the problem
of structural induction. The terminals in the grammar formalism denote
the set of observables in our induction problem. The NonTerminals
stand for internal variables in terms of which the observations (terminals)
are explained. The {}``explanation'' is given by a derivation of
the stream of observations from the initial symbol $S\overset{*}{\to}w$.
The NonTerminals that appear in the derivation are the internal variables
in terms of which the surface structure given by the stream of observations
$w$ is explained. Given this correspondence, structural induction
aims to find an appropriate set of NonTerminals $N$ and a set of
rewrite rules $R$ that will allow us to derive (explain) the input
stream of observations $w$ from the initial symbol $S$. The process
of Structural Induction may invent a new rewrite rule $l\to r$ under
certain conditions and this new rule may contain in turn new NonTerminals
(internal variables) which are added to the already existing ones.
The common intuition is that $l$ is a simpler version of $r$, as
the final goal is to reduce $w$ to $S$. The terminals constitute
the input symbols (standing for observables), the NonTerminals constitute
whatever additional {}``internal'' variables that are needed, the
rewrite rules describe their interrelationship and altogether they
constitute the ontology. The correspondence between the terms used
in structural induction and generative grammars is summarized in Table
\ref{tab:Correspondence-between-Structural}.

\begin{table}
\begin{centering}
\begin{tabular}{|c|c|}
\hline 
Structural Induction & Generative Grammar\tabularnewline
\hline
\hline 
Observables & Terminals $T$\tabularnewline
\hline 
Internal Variables & NonTerminals $N$\tabularnewline
\hline 
Law / Theory & production rule(s) $l\to r$\tabularnewline
\hline 
Ontology & Grammar $G$\tabularnewline
\hline 
Observations Stream  & word $w$\tabularnewline
\hline 
Explanation & Derivation $S\overset{*}{\to}w$\tabularnewline
\hline 
Partial Explanation & Derivation $\alpha\overset{*}{\to}w$\tabularnewline
\hline
\end{tabular}
\par\end{centering}

\caption{\label{tab:Correspondence-between-Structural}Correspondence between
Structural Induction and Generative Grammars }
\end{table}

Thus, in general, structural induction may invent any rewrite rule
of the form $l\to r$, potentially introducing new NonTerminals, the
problem is that there are infinitely many such rules that we could
invent at any point in time. In order to make the process more well
defined we ask whether it is possible to find a set of fundamental
structural operations which is finite and minimalistic, such that
all the rules (or more precisely sets of rules) can be expressed in
terms of these operations. This would establish a normal form in terms
of a finite set of operations and then the problem of generating laws
will be reduced to making appropriate choices from this set without
sacrificing completeness. In the next subsection we will attempt to
decompose the rules $l\to r$ into a small finite set of fundamental
structural elements which will allow us to design better structure
search mechanisms. 

\vspace*{-0.1in}

\subsection{ASNF Theorems}

\textbf{Issue} ($\epsilon-Construction$). In the rest of the paper
we will prove some theorems that impose various sets of conditions
on a grammar $G$ in order for the grammar to be considered in a certain
Normal Form. If \emph{$\epsilon\in L(G)$ }however,\emph{ }we will
allow two specific rules of the grammar $G$ to be exempted from these
constraints and still consider the grammar in the Normal Form. More
exactly if \emph{$\epsilon\in L(G)$} and given a grammar $G'$ such
that $L(G')=L(G\backslash\{\epsilon\})$ and $G'=(N',T,S',R')$ is
in a certain Normal Form then the grammar $G=(N\cup\{S\},T,S,R=R'\cup\{S\to\epsilon,S\to S'\})$
where $S\notin N'$ will also be considered in that certain Normal
Form despite the fact that the two productions $\{S\to\epsilon,S\to S'\}$
may violate the conditions of the Normal Form. These are the only
productions that will be allowed to violate the Normal Form conditions.
Note that $S$ is a brand new NonTerminal and does not appear in any
other productions aside from these two. Without loss of generality
we will assume in the rest of the paper that $\epsilon\notin L(G)$.
This is because if $\epsilon\in L(G)$ we can always produce using
the above-mentioned construction a grammar $G''$ that is in a certain
Normal Form and $L(G'')=L(G')$ from a grammar $G'$ that is in that
Normal Form and satisfies $L(G')=L(G\backslash\{\epsilon\})$. We
will call the procedure just outlined the $\epsilon-Construction$.
We will call the following statement the $\epsilon-Amendment$: Let
\emph{$G=(N,T,S,R)$ }be a grammar, if $\epsilon$ is an element of
the language $L(G)$ one rule of the form $S\to\epsilon$ is allowed
and furthermore the restriction that $S$ does not appear in the right
hand side of any rule is imposed. 

First we state a weak form of the Abstraction SuperStructuring Normal
Form for Context Free Grammars. 

\textbf{Theorem 2 (Weak-CFG-ASNF).} \emph{Let $G=(N,T,S,R)$, }$\epsilon\notin L(G)$\emph{
be a Context Free Grammar. Then there exists a Context Free Grammar
$G'$ such that $L(G)=L(G')$ and $G'$ contains only rules of the
following type:}
\begin{enumerate}
\item \emph{$A\to B$}
\item \emph{$A\to BC$ }
\item \emph{$A\to a$ }
\end{enumerate}
\emph{Proof . }Since $G$ is a CFG it can be written in the Chomsky
Normal Form \cite{Salomaa 1985}. That is, such that it contains only
productions of the forms 2 and 3. If $\epsilon\in L(G)$ a rule of
the form $S\to\epsilon$ is allowed and $S$ does not appear in the
rhs of any other rule\emph{ }($\epsilon-Amendment$). Since we have
assumed that $\epsilon\notin L(G)$ we do not need to deal with $\epsilon-Amendment$
and hence the proof. 

$\Box$

\textbf{Remarks. }
\begin{enumerate}
\item We will call the rules of type 1 Renamings (REN).
\item We will call the rules of type 2 SuperStructures (SS) or compositions.
\item The rules of the type 3 are just convenience renamings of observables
into internal variables in order to uniformize the notation and we
will call them Terminal (TERMINAL). 
\end{enumerate}
We are now ready to state the the Weak ASNF theorem for the general
case.

\textbf{Theorem 3 (Weak-GEN-ASNF).} \emph{Let $G=(N,T,S,R)$, }$\epsilon\notin L(G)$\emph{
be a General (unrestricted) Grammar. Then there exists a grammar $G'$
such that $L(G)=L(G')$ and $G'$ contains only rules of the following
type:}
\begin{enumerate}
\item \emph{$A\to B$}
\item \emph{$A\to BC$ }
\item \emph{$A\to a$ }
\item \emph{$AB\to C$ }
\end{enumerate}
\emph{Proof .} See Appendix.

\textbf{Remark. }We will call the rules of type 4 Reverse Super-Structuring
(RSS).

In the next theorem we will strengthen our results by allowing only
the renamings (REN) to be non unique. First we define what we mean
by uniqueness and then we proceed to state and prove a lemma that
will allow us to strengthen the Weak-GEN-ASNF by imposing uniqueness
on all the productions safe renamings.

\textbf{Definition (}\textbf{\emph{strong-uniqueness}}\textbf{).}
We will say that a production $\alpha\to\beta$ respects \emph{strong-uniqueness}
if this is the only production that has the property that it has $\alpha$
in the lhs and also this is the only production that has $\beta$
on the rhs.

\textbf{Lemma 2. }\emph{Let $G=(N,T,S,R)$, $\epsilon\notin G$ a
grammar such that all its productions are of the form:}
\begin{enumerate}
\item $A\to B$
\item $A\to\zeta$ , $\zeta\notin N$
\item $\zeta\to B$ , $\zeta\notin N$
\end{enumerate}
\emph{Modify the the grammar G to obtain $G'=(N',T,S',R')$ as follows:}
\begin{enumerate}
\item \emph{Introduce a new start symbol $S'$ and the production $S'\to S$.}
\item \emph{For each $\zeta\notin N$ that appears in the rhs of one production
in $G$ let $\{A_{i}\to\zeta\}_{i=1,n}$ all the the productions that
contain $\zeta$ in the rhs of a production. Introduce a new NonTerminal
$X_{\zeta}$ and the productions $X_{\zeta}\to\zeta$ and $\{A_{i}\to X_{\zeta}\}_{i=1,n}$
and eliminate the old productions $\{A_{i}\to\zeta\}_{i=1,n}$.}
\item \emph{For each $\zeta\notin N$ that appears in the lhs of one production
in $G$ let $\{\zeta\to B_{j}\}_{j=1,m}$ all the the productions
that contain $\zeta$ the lhs of a production. Introduce a new NonTerminal
$Y_{\zeta}$ and the productions $\zeta\to Y_{\zeta}$ and $\{Y_{\zeta}\to B_{j}\}_{j=1,m}$
and eliminate the old productions $\{\zeta\to B_{j}\}_{j=1,m}$.}
\end{enumerate}
\emph{Then the new grammar $G'$ generates the same language as the
initial grammar $G$ and all the productions of the form $A\to\zeta$
and $\zeta\to B$ , $\zeta\notin N$ respect strong-uniqueness. Furthermore,
if the initial grammar has some restrictions on the composition of
the $\zeta\notin N$ that appears in the productions of type 2 and
3, they are respected since $\zeta$ is left unchanged in the productions
of the new grammar and the only other types of productions introduced
are renamings that are of neither type 2 nor type 3. }

\emph{Proof}. See Appendix \cite{Silvescu 2011}.

By applying Lemma 2 to the previous two Weak-ASNF theorems we obtain
strong versions of these theorems which enforce \emph{strong-uniqueness}
in all the productions safe the renamings.

\textbf{Theorem 4 (Strong-CFG-ASNF).} \emph{Let $G=(N,T,S,R)$, }$\epsilon\notin L(G)$\emph{
be a Context Free Grammar. Then there exists a Context Free Grammar
$G'$ such that $L(G)=L(G')$ and $G'$ contains only rules of the
following type:}
\begin{enumerate}
\item \emph{$A\to B$}
\item \emph{$A\to BC$ - and this is the only rule that has $BC$ in the
rhs and this is the only rule that has $A$ in the lhs (strong-uniqueness).}
\item \emph{$A\to a$ - and this is the only rule that has $a$ in the rhs
and this is the only rule that has $A$ in the lhs (strong-uniqueness).}
\end{enumerate}
\emph{Proof}. Apply Lemma 2 to the grammar converted into Weak-CFG-ASNF$\Box$

\textbf{Theorem 5 (Strong-GEN-ASNF).} \emph{Let $G=(N,T,S,R)$, }$\epsilon\notin L(G)$\emph{
be a general (unrestricted) grammar. Then there exists a grammar $G'$
such that $L(G)=L(G')$ and $G'$ contains only rules of the following
type:}
\begin{enumerate}
\item \emph{$A\to B$}
\item \emph{$A\to BC$ - and this is the only rule that has $BC$ in the
rhs and this is the only rule that has $A$ in the lhs (strong-uniqueness).}
\item \emph{$A\to a$ - and this is the only rule that has $a$ in the rhs
and this is the only rule that has $A$ in the lhs (strong-uniqueness).}
\item \emph{$AB\to C$ - and this is the only rule that has $C$ in the
rhs and this is the only rule that has $AB$ in the lhs (strong-uniqueness).}
\end{enumerate}
\emph{Proof}. Apply Lemma 2 to the grammar converted into Weak-GEN-ASNF$\Box$

\textbf{Remark. }After enforcing strong uniqueness the only productions
that contain choice are those of type 1 - renamings (REN). 

In the light of this theorem we proceed to introduce the concept of
abstraction and prove some additional results. 

\vspace*{-0.1in}

\subsection{Abstractions And Reverse Abstractions}

\textbf{Definitions (Abstractions Graph).} Given a grammar $G=(N,T,S,R)$
which is in an ASNF from any of the Theorems 1 - 4 we call an \emph{Abstractions
Graph of the grammar $G$ }and denote it by $AG(G)$ a Directed Graph
$G=(N,E)$ whose nodes are the NonTerminals of the grammar $G$ and
whose edges are constructed as follows: we put a directed edge starting
from $A$ and ending in $B$ iff $A\to B$ is a production that occurs
in the grammar. Without loss of generality, we can assume that the
graph has no self loops, i.e., edges of the form $A\to A$; If such
self-loops exist, the corresponding productions can be eliminated
from the grammar without altering the language. In such a directed
graph a node $A$ has a set of outgoing edges and a set of incoming
edges which we refer to as out-edges and in-edges respectively. We
will call a node $A$ along with its out-edges the \emph{Abstraction
at A} and denote it $ABS(A)=\{A,OE_{A}=\{(A,B)|(A,B)\in E\}\}$. Similarly,
we will call a node $A$ along with its in-edges the \emph{Reverse
Abstraction at A} and denote it $RABS(A)=\{A,IE_{A}=\{(B,A)|(B,A)\in E\}\}$.

\vspace*{-0.1in}

\subsection{Grow Shrink Theorem}

\textbf{Theorem 6. }\emph{Let $G=(N,T,S,R)$, }$\epsilon\notin L(G)$\emph{
be a General Grammar. Then we can convert such a grammar into the
Strong-GEN-ASNF i.e., such that all the productions are of the following
form: }
\begin{enumerate}
\item \emph{$A\to B$}
\item \emph{$A\to BC$ - and this is the only rule that has $BC$ in the
rhs and this is the only rule that has $A$ in the lhs. (strong-uniqueness)}
\item \emph{$A\to a$ - and this is the only rule that has $A$ on the lhs
and there is no other rule that has $a$ on the rhs. (strong uniqueness) }
\item \emph{$AB\to C$ - and this is the only rule that has $C$ in the
rhs and this is the only rule that has $AB$ in the lhs. (strong-uniqueness)}
\end{enumerate}
\emph{And furthermore for any derivation $w$ such that $\gamma\overset{*}{\to}w$
, in $G$, $\gamma\in N^{+}$ there exists a derivation $\gamma\overset{*}{\to}\mu\overset{*}{\to}\nu\overset{*}{\to}w$
such that $\mu\in N^{+}$, $\nu\in N^{*}$ and $\gamma\overset{*}{\to}\mu$
contains only rules of type 1 and 2 (REN, SS), $\mu\overset{*}{\to}\alpha$
contains only rules of the type 1, more particularly only Reverse
Abstractions and type 4 (REN(RABS), RSS) and $\nu\overset{*}{\to}w$
contains only rules of type 3 (TERMINAL).}

\emph{Proof}. See Appendix.

We have therefore proved that for each General Grammar $G$ we can
transform it in a Strong-GEN-ASNF such that the derivation (explanation
in structural induction terminology) of any terminal string $w$ can
be organized in three phases such that: Phase 1 uses only productions
that grow (or leave unchanged) the size of the intermediate string;
Phase 2 uses only productions that shrink (or leave unchanged) the
size of the intermediate string; and Phase 3 uses only TERMINAL productions%
\footnote{At first sight, it may seem that this construction offers a way to
solve the halting problem. However, this is not the case, since we
do not answer the question of deciding when to stop expanding the
current string and start shrinking which is key to solving the halting
problem. %
}. In the case of grammars that are not in the normal form as defined
above, the situation is a little more complicated because of successive
applications of grow and shrink phases. However, we have shown that
we can always transform an arbitrary grammar into one that in the
normal form. Note further that the grow phase in both theorems use
only context free productions.

We now proceed to examine the implications of the preceeding results
in the larger context including the nature of hidden variables, radical
positivism and the David Hume's principles of \textit{connexion} among
ideas. 

\vspace*{-0.1in}

\section{The Fundamental Operations of Structural Induction }

Recall that our notion of structural induction entails: Given a sequence
of observations $w$ we attempt to find a theory (grammar) that explains
$w$ and simultaneously also the explanation (derivation) $S\overset{*}{\to}w$.
In a local way we may think that whenever we have a production rule
$l\to r$ that $l$ explains $r$. In a bottom up - data driven way
we may proceed as follows: First introduce for every observable $a$
a production $A\to a$. The role of these productions is simply to
bring the observables into the realm of internal variables. The resulting
association is between the observables and the corresponding internal
variables unique (one to one and onto) and hence, once this association
is established, we can forget about the existence of bbservables (Terminals).
Since establishing these associations is the only role of the TERMINAL
productions, they are not true structural operations. With this in
mind, if we are to construct a theory in the GEN-ASNF we can postulate
laws of the following form:
\begin{enumerate}
\item $A\to BC$ - Super-structuring (SS) which takes two internal variables
$B$ and $C$ that occur within proximity of each other (adjacent)
and labels the compound. Henceforth, the shorter name $A$ can be
used instead for $BC$ . This is the sole role of super-structuring
- to give a name to a composite structure to facilitate shorter explanations
at latter stages.
\item $A\to B|C$ - Abstraction (ABS). Introduces a name for the occurrence
of either of the variables $B$ or $C$. This allows for compactly
representing two productions that are identical except that one uses
$B$ and the uses $C$ by a single production using $A$. The role
of Abstraction is to give a name to a group of entities (we have chosen
two only for simplicity) in order to facilitate more general explanations
at latter stages which in turn will produce more compact theories.
\item $AB\to C$ - Reverse Super-structuring (RSS) which introduces up to
two existing or new internal variables that are close to each other
(with respect to a specified topology) that together {}``explain''
the internal variable $C$.
\item $A\to C$, $B\to C$ - Reverse Abstraction (RABS) which uses existing
or new internal variables \textit{$A$} and $B$ as alternative explanations
of the internal variable $C$ (we have chosen two variables only for
simplicity). 
\end{enumerate}
\vspace*{-0.1in}

\subsection{Reasons for Postulating Hidden Variables}

Recall that are at least two types of reasons for creating Hidden
Variables:
\begin{enumerate}
\item (\textbf{OR} type) - {[}multiple alternative hidden causes{]} The
OR type corresponds to the case when some visible effect can have
multiple hidden causes $H1\to Effect$, $H2\to Effect$ . In our setting,
this case corresponds to Reverse Abstraction. One typical example
of this is: The grass is wet, and hence either it rained last night
or the sprinkler was on. In the statistical and machine learning literature
the models that use this type of hidden variables are called mixture
models \cite{Lauritzen 1996 5}.
\item (\textbf{T-AND} type) - {[}multiple concurrent hidden causes{]} The
T-AND type, i.e., topological AND type, of which the AND is a sepcial
case. This corresponds to the case when one visible effect has two
hidden causes both of which have to occur within proximity of each
other (with respect to a specified topology) in order to produce the
visible effect. $H1H2\to Effect$. In our setting, this corresponds
Reverse Super-structuring. In the Statistical / Graphical Models literature
the particular case of AND hidden explanations is the one that introduces
edges between hidden variables in the depedence graph \cite{Elidan and Friedman 2005},
\cite{Lauritzen 1996 5}, \cite{Pearl 1988 5}.
\end{enumerate}
The preceeding discussion shows that we can associate with two possible
reasons for creating hidden variables, the structural operations of
Reverse Abstraction and Reverse Super Structuring respectively. Because
these are the only two types of productions that introduce hidden
variables in the GEN-ASNF this provides a characterization of the
rationales for introducing hidden variables. 

\vspace*{-0.1in}

\subsection{Radical Positivism}

If we rule out the use of RSS and RABS, the only operations that involve
the postulation of hidden variables, we are left with only SS and
ABS which corresponds to the radical positivist \cite{Ayer 1936}
stance under the computationalist assumption. An explanation of a
stream of observations $w$ in the Radical Positivist theory of the
world is mainly a theory of how the observables in the world are grouped
into classes (Abstractions) and how smaller chunks of observations
are tied together into bigger ones (Super-Structures). The laws of
the radical positivist theory are truly empirical laws as they only
address relations among observations. 

However, structural induction, if it is constrained to using only
ABS and SS, the class of theories that can be induced is necessarily
a subset of the set of theories that can be described by Turing Machines.
More precisely, the resulting grammars will be a strict subset of
Context Free Grammars, (since CFG contain SS, and REN(ABS+RABS)).
Next we will examine how any theory of the world may look like from
the most general perspective when we do allow Hidden Variables. 

\vspace*{-0.1in}

\subsection{General theories of the world}

If structural induction is allowed to take advantage of RSS and RABS
in addition to SS and ABS, the resulting theories can make use of
hidden variables. Observations are a derivative byproduct obtained
from a richer hidden variable state description by a reduction: either
of size - performed by Reverse SuperStructuring or of information
- performed by Reverse Abstraction. Note that, while in general, structural
induction can alternate several times between REN+SS and RABS+RSS,
we have shown that three phases suffice: a growth phase (REN+SS);
a shrink phase (RABS+RSS); and a Terminal phase. Whether we can push
all the RABS from the first phase into the second phase and make the
first phase look like the one in the radical positivist stance (only
ABS+SS) remains an open question (See Appendix  for a Conjecture to
this effect). 

\vspace*{-0.1in}

\subsection{Hume's principles of connexion among ideas}

We now examine, against the backdrop of GEN-ASNF theorem, a statement
made by philosopher David Hume more that 2 centuries ago: \emph{{}``I
do not find that any philosopher has attempted to enumerate or class
all the principles of association {[}of ideas{]}. ... To me, there
appear to be only three principles of connexion among ideas, namely,
Resemblance, Contiguity in time or place, and Cause and Effect''
\cite{Hume 1993}}. If we substitute Resemblance with Abstraction
(since abstraction is triggered by resemblance or similarity), Contiguity
in time or place with Super-Structuring (since proximity, e.g., spatio-temporal
proximity drives Super-Structuring) and Cause and Effect with the
two types of explanations that utilize hidden variables, it is easy
to see that the GEN-ASNF theorem is simply a precise restatement of
Hume's claim under the computationalist assumption. 

\vspace*{-0.1in}

\section{Summary}

We have shown that \textit{abstraction} (grouping \textit{similar}
entities) and super-structuring (combining topologically e.g., spatio-temporally
close entities) as the essential structural operations in the induction
process. A structural induction process that relies only on abstraction
and super-structuring corresponds to the radical positivist stance.
We have shown that only two more structural operations, namely, \textit{reverse
abstraction} and \textit{reverse super-structuring} (the duals of
abstraction and super-structuring respectively) (a) suffice in order
to exploit the full power of Turing-equivalent generative grammars
in induction; and (b) operationalize two rationales for the introduction
of hidden variables into theories of the world. The GEN-ASNF theorem
can be seen as simply a restatement, under the computationalist assumption,
of Hume's 2-century old claim regarding the principles of connexion
among ideas.

\pagebreak{}

\section*{Appendix }

\textbf{Theorem 3 (Weak-GEN-ASNF).} \emph{Let $G=(N,T,S,R)$, }$\epsilon\notin L(G)$\emph{
be a General (unrestricted) Grammar. Then there exists a grammar $G'$
such that $L(G)=L(G')$ and $G'$ contains only rules of the following
type:}
\begin{enumerate}
\item \emph{$A\to B$}
\item \emph{$A\to BC$ }
\item \emph{$A\to a$ }
\item \emph{$AB\to C$ }
\end{enumerate}
\emph{Proof .} Every general grammar can be written in the Generalized
Kuroda Normal Form(GKNF) \cite{Salomaa 1985}. That is, it contains
only productions of the form: 
\begin{enumerate}
\item $A\to\epsilon$
\item $AB\to CD$
\item $A\to BC$
\item $A\to a$ 
\end{enumerate}
Assume that we have already converted our grammar in the GKNF. We
will prove our claim in 3 steps.

\textbf{\underbar{Step 1}}. For each $\{AB\to CD\}$ we introduce
a new NonTerminal $X_{AB,CD}$ and the following production rules
$\{AB\to X_{AB,CD}\}$ and $\{X_{AB,CD}\to CD\}$ and eliminate the
old $\{AB\to CD\}$ production rules. In this way we ensure that all
the rules of type 2 in GKNF have been rewritten into rules of types
2 and 4 in the GEN-ASNF. 

The new grammar generates the same language. To see this let $oldG=(N_{oldG},T,S,R_{oldG})$
denote the old grammar and $newG=(N_{newG},T,S,R_{newG})$ denote
the new grammar. Let \emph{$\gamma\in N_{oldG}^{+}$} and $\gamma\overset{*}{\to}w$
be a derivation in\emph{ $oldG$}. In this derivation we can replace
all the uses of the production $AB\to CD$ with the sequence $AB\to X_{AB,CD}\to CD$
and get a derivation that is valid in $newG$, since all the other
productions are common between the two grammars. Thus we have proved
that for all \emph{$\gamma\in N_{oldG}^{+}$} we can convert a valid
derivation from $oldG$, $\gamma\overset{*}{\to}w$ into a valid derivation
in $newG$ and in particular this is true also for $S\overset{*}{\to}w$.
Therefore $L(oldG)\subseteq L(newG)$. Conversely, let \emph{$\gamma\in N_{oldG}^{+}$}
and $\gamma\overset{*}{\to}w$ be a valid derivation in $newG$ then
whenever we use the rule $AB\to X_{AB,CD}$ in this derivation $\gamma\overset{*}{\to}\alpha AB\beta\to\alpha X_{AB,CD}\beta\overset{*}{\to}w$
let $\gamma\overset{*}{\to}\alpha AB\beta\to\alpha X_{AB,CD}\beta\overset{*}{\to}\delta X_{AB,CD}\eta\to\delta CD\eta\overset{*}{\to}w$
be the place where the $X_{AB,CD}$ that occurs between $\alpha$
and $\beta$ is rewritten (used in the lhs of a production, even if
it rewrites to the same symbol) for the first time. Then necessarily
the $X_{AB,CD}\to CD$ rule is the one that applies since it is the
only one that has $X_{AB,CD}$ in the lhs. Furthermore, as a consequence
of Lemma 1 we have that $\alpha\overset{*}{\to}\delta$ and $\beta\overset{*}{\to}\eta$
are valid derivations in $newG$. Therefore we can bring the production
$X_{AB,CD}\to CD$ right before the use of $AB\to X_{AB,CD}$ as follows
$\gamma\overset{*}{\to}\alpha AB\beta\to\alpha X_{AB,CD}\beta\to\alpha CD\beta\overset{*}{\to}\delta CD\beta\overset{*}{\to}\delta CD\eta\overset{*}{\to}w$
and still have a valid derivation in $newG$. We can repeat this procedure
for all places where rules of the form $AB\to X_{AB,CD}$ appear.
In this modified derivation we can replace all the uses of the sequence
$AB\to X_{AB,CD}\to CD$ with the production $AB\to CD$ and obtain
a derivation that is valid in $oldG$ since all the other productions
are common between the two grammars. Thus we have proved that for
all \emph{$\gamma\in N_{oldG}^{+}$} we can convert a valid derivation
from $newG$ ,$\gamma\overset{*}{\to}w$ into a valid derivation in
$oldG$ and in particular this is true also for $S\overset{*}{\to}w$
since $S\in N_{oldG}^{+}$, and therefore $L(newG)\subseteq L(oldG)$.
Therefore we have proved that the grammars are equivalent, i.e., $L(oldG)=L(newG)$.

\textbf{\underbar{Step 2}}. Returning to the main argument, so far
our grammar has only rules from Weak-GEN-ASNF safe for the $\epsilon$-productions
$A\to\epsilon$. Next we will eliminate $\epsilon$-productions $A\to\epsilon$
in two steps. First for each $A\to\epsilon$ we introduce a new NonTerminal
$X_{A,E}$ and the productions $X_{A,E}\to E$ and $E\to\epsilon$
and eliminate $A\to\epsilon$, where $E$ is a distinguished new NonTerminal
(that will basically stand for $\epsilon$ internally). This insures
that we have only one $\epsilon$-production, namely $E\to\epsilon$
and $E$ does not appear on the lhs of any other production and also
that all the rules that rewrite to $E$ are of the form $A\to E$. 

The new grammar generates the same language. We will use a similar
proof technique as in the previous step. Let $oldG=(N_{oldG},T,S,R_{oldG})$
denote the old grammar and $newG=(N_{newG},T,S,R_{newG})$ denote
the new grammar. Let \emph{$\gamma\in N_{oldG}^{+}$} and $\gamma\overset{*}{\to}w$
be a derivation in\emph{ $oldG$}. In this derivation we can replace
all the uses of the production $A\to\epsilon$ with the sequence $X_{A,E}\to E\to\epsilon$
and get a derivation that is valid in $newG$, since all the other
productions are common between the two grammars. Thus we have proved
that for all \emph{$\gamma\in N_{oldG}^{+}$} we can convert a valid
derivation from $oldG$ ,$\gamma\overset{*}{\to}w$ into a valid derivation
in $newG$ and in particular this is true also for $S\overset{*}{\to}w$,
therefore $L(oldG)\subseteq L(newG)$. Conversely, let \emph{$\gamma\in N_{oldG}^{+}$}
and $\gamma\overset{*}{\to}w$ be a valid derivation in $newG$ then
whenever we use the rule $X_{A,E}\to E$ in this derivation $\gamma\overset{*}{\to}\alpha X_{A,E}\beta\to\alpha E\beta\overset{*}{\to}w$
let $\gamma\overset{*}{\to}\alpha X_{A,E}\beta\to\alpha E\beta\overset{*}{\to}\delta E\eta\to\delta\eta\overset{*}{\to}w$
be the place where the $E$ that occurs between $\alpha$ and $\beta$
is rewritten(used in the lhs of a production, even if it rewrites
to the same symbol) for the first time. Then necessarily the $E\to\epsilon$
rule is the one that applies since it is the only one that has $E$
in the lhs. Furthermore, as consequence of Lemma 1 we have that $\alpha\overset{*}{\to}\delta$
and $\beta\overset{*}{\to}\eta$ are valid derivations in $newG$.
Therefore we can bring the production $E\to\epsilon$ right before
the use of $X_{A,E}\to E$ as follows $\gamma\overset{*}{\to}\alpha X_{A,E}\beta\to\alpha E\beta\to\alpha\beta\overset{*}{\to}\delta\beta\overset{*}{\to}\delta\eta\overset{*}{\to}w$
and still have a valid derivations in $newG$. We can repeat this
procedure for all places where rules of the form $X_{A,E}\to E$ appear.
In this modified derivation we can replace all the uses of the sequence
$X_{A,E}\to E\to\epsilon$ with the production $A\to\epsilon$ and
obtain a derivation that is valid in $oldG$ since all the other productions
are common between the two grammars. Thus we have proved that for
all \emph{$\gamma\in N_{oldG}^{+}$} we can convert a valid derivation
from $newG$, $\gamma\overset{*}{\to}w$ into a valid derivation in
$oldG$ and in particular this is true also for $S\overset{*}{\to}w$
since $S\in N_{oldG}^{+}$, and therefore $L(newG)\subseteq L(oldG)$.
Therefore we have proved that the grammars are equivalent, i.e., $L(oldG)=L(newG)$.

\textbf{\underbar{Step 3}}. To summarize: the new grammar has only
rules of the Weak-GEN-ASNF type, safe for the production $E\to\epsilon$
which is the only production that has $E$ in the lhs\emph{ }and there
is no other rule that has $\epsilon$ on the rhs (\emph{strong-uniqueness})
and furthermore the only rules that contain $E$ in the rhs are of
the form\emph{ $A\to E$ }(\emph{only renamings to E}). 

We will eliminate the $\epsilon$-production $E\to\epsilon$ as follows:
Let $\{A_{i}\to E\}_{i=1,n}$ be all the productions that have \emph{$E$}
on the rhs. For all NonTerminals $X\in N\cup T$ introduce productions
$\{XA_{i}\to X\}_{i=1,n}$ and $\{A_{i}X\to X\}_{i=1,n}$ and eliminate
$\{A_{i}\to E\}_{i=1,n}$, furthermore we also eliminate $E\to\epsilon$.

The new grammar generates the same language. We will use a similar
proof technique as in the previous step. Let $oldG=(N_{oldG},T,S,R_{oldG})$
denote the old grammar and $newG=(N_{newG},T,S,R_{newG})$ denote
the new grammar. Let \emph{$\gamma\in N_{oldG}^{+}$} and $\gamma\overset{*}{\to}w$
be a derivation in\emph{ $oldG$}. Then whenever we use a rule of
the form $A_{i}\to E$ in this derivation $\gamma\overset{*}{\to}\alpha A_{i}\beta\to\alpha E\beta\overset{*}{\to}w$
let $\gamma\overset{*}{\to}\alpha A_{i}\beta\to\alpha E\beta\overset{*}{\to}\delta E\eta\to\delta\eta\overset{*}{\to}w$
be the place where the $E$ that occurs between $\alpha$ and $\beta$
is rewritten(used in the lhs of a production, even if it rewrites
to the same symbol) for the first time. Then necessarily the $E\to\epsilon$
rule is the one that applies since it is the only one that has $E$
in the lhs. Furthermore, as a consequence of Lemma 1 we have that
$\alpha\overset{*}{\to}\delta$ and $\beta\overset{*}{\to}\eta$ are
valid derivations in $oldG$. Therefore we can bring the production
$E\to\epsilon$ right before the use of $A_{i}\to E$ as follows $\gamma\overset{*}{\to}\alpha A_{i}\beta\to\alpha E\beta\to\alpha\beta\overset{*}{\to}\delta\beta\overset{*}{\to}\delta\eta\overset{*}{\to}w$
and still have a valid derivations in $oldG$. We can repeat this
procedure for all places where rules of the form $A_{i}\to E$ appear.
In this modified derivation we can replace all the uses of the sequence
$A_{i}\to E\to\epsilon$ with the production $XA_{i}\to X$ if $\alpha\neq\epsilon$
and $XA_{i}\to X$ if $\beta\neq\epsilon$ and obtain a derivation
that is valid in $newG$ since all the other productions are common
between the two grammars, (e.g., if $\alpha\neq\epsilon$, $\alpha=\alpha_{1}X$
replace $\alpha A_{i}\beta=\alpha_{1}XA_{i}\beta\to\alpha_{1}XE\beta\to\alpha_{1}X\beta=\alpha\beta$
which is valid in $oldG$ with $\alpha A_{i}\beta=\alpha_{1}XA_{i}\beta\to\alpha_{1}X\beta=\alpha\beta$
which is valid in $newG$ and similarly for $\beta\neq\epsilon$).
In this way since all the other productions are common between the
two grammars we can convert a derivation $\gamma\overset{*}{\to_{oldG}}w$
into a derivation $\gamma\overset{*}{\to_{newG}}w$. Note that it
is not possible for both $\alpha$ and $\beta$ to be equal to $\epsilon$
because this will imply that the derivation $\gamma\overset{*}{\to}w$
is $\gamma\overset{*}{\to}A_{i}\to E\to\epsilon$ which contradicts
the hypothesis that $w\in T^{+}$. Thus we have proved that for all
\emph{$\gamma\in N_{oldG}^{+}$} we can convert a valid derivation
from $oldG$, $\gamma\overset{*}{\to}w$ into a valid derivation in
$newG$ and in particular this is true also for $S\overset{*}{\to}w$
since $S\in N_{oldG}^{+}$, and therefore $L(oldG)\subseteq L(newG)$.

Conversely, let \emph{$\gamma\in N_{oldG}^{+}$} and $\gamma\overset{*}{\to}w$
be a derivation in\emph{ $newG$} . In this derivation we can replace
all the uses of the production $XA_{i}\to X$ with the sequence $A_{i}\to E\to\epsilon$
and get a derivation that is valid in $oldG$ since all the other
productions are common between the two grammars (e.g., we replace
$\alpha A_{i}\beta=\alpha_{1}XA_{i}\beta\to\alpha_{1}X\beta=\alpha\beta$
which is valid in $newG$ with $\alpha A_{i}\beta=\alpha_{1}XA_{i}\beta\to\alpha_{1}XE\beta\to\alpha_{1}X\beta=\alpha\beta$
which is valid in $oldG$). We proceed similarly with the productions
of the form $A_{i}X\to X$ and replace them with the sequence $A_{i}\to E\to\epsilon$
and get a derivation that is valid in $oldG$. Thus we have proved
that for all \emph{$\gamma\in N_{oldG}^{+}$} we can convert a valid
derivation from $newG$, $\gamma\overset{*}{\to}w$ into a valid derivation
in $oldG$ and in particular this is true also for $S\overset{*}{\to}w$,
therefore $L(newG)\subseteq L(oldG)$. Hence we have proved that the
grammars are equivalent, i.e., $L(oldG)=L(newG)$.

$\Box$

\textbf{Lemma 1.} \emph{Let $G=(N,T,S,R)$ be a grammar and let $\alpha\mu\beta\overset{*}{\to}\delta\mu\eta$,
$\mu\in(N\cup T)^{+}$ a valid derivation in $G$ that does not rewrites
the $\mu$ (uses productions whose lhs match any part of $\mu$ even
if it rewrites to itself), occurring between $\alpha$ and $\beta$,
then }$\alpha\overset{*}{\to}\delta$, $\beta\overset{*}{\to}\eta$
\emph{are valid derivations in $G$.}

\emph{Proof. }Because $\alpha\mu\beta\overset{*}{\to}\delta\mu\eta$
does not rewrites any part of $\mu$ (even to itself) it follows that
the lhs of any production rule that is used in this derivation either
matches a string to the left of $\mu$ or to the right of $\mu$.
If we start from $\alpha$ and use the productions that match to the
left in the same order as in $\alpha X\beta\overset{*}{\to}\delta X\eta$
then we necessarily get $\alpha\overset{*}{\to}\delta$, a valid derivation
in $G$. Similarly, by using the productions that match to the right
of $\mu$ we get $\beta\overset{*}{\to}\eta$ valid in $G$.

$\Box$

\textbf{Lemma 2. }\emph{Let $G=(N,T,S,R)$, $\epsilon\notin G$ a
grammar such that all its productions are of the form:}
\begin{enumerate}
\item $A\to B$
\item $A\to\zeta$ , $\zeta\notin N$
\item $\zeta\to B$ , $\zeta\notin N$
\end{enumerate}
\emph{Modify the the grammar into a new grammar $G'=(N',T,S',R')$obtained
as follows:}
\begin{enumerate}
\item \emph{Introduce a new start symbol $S'$ and the production $S'\to S$.}
\item \emph{For each $\zeta\notin N$ that appears in the rhs of one production
in $G$ let $\{A_{i}\to\zeta\}_{i=1,n}$ all the the productions that
contain $\zeta$ in the rhs of a production. Introduce a new NonTerminal
$X_{\zeta}$ and the productions $X_{\zeta}\to\zeta$ and $\{A_{i}\to X_{\zeta}\}_{i=1,n}$
and eliminate the old productions $\{A_{i}\to\zeta\}_{i=1,n}$.}
\item \emph{For each $\zeta\notin N$ that appears in the lhs of one production
in $G$ let $\{\zeta\to B_{j}\}_{j=1,m}$ all the the productions
that contain $\zeta$ the lhs of a production. Introduce a new NonTerminal
$Y_{\zeta}$ and the productions $\zeta\to Y_{\zeta}$ and $\{Y_{\zeta}\to B_{j}\}_{j=1,m}$
and eliminate the old productions $\{\zeta\to B_{j}\}_{j=1,m}$.}
\end{enumerate}
\emph{Then the new grammar $G'$ generates the same language as the
initial grammar $G$ and all the productions of the form $A\to\zeta$
and $\zeta\to B$ , $\zeta\notin N$ respect strong-uniqueness. Furthermore
if the initial grammar has some restrictions on the composition of
the $\zeta\notin N$ that appears in the productions of type 2 and
3, they are still maintained since $\zeta$ is left unchanged in the
productions of the new grammar and the only other types of productions
introduced are Renamings which do not belong to either type 2 or 3. }

\emph{Proof}. To show that the two grammars are equivalent we will
use a similar proof technique as in the previous theorem. Let \emph{$\gamma\in N^{+}$}
and $\gamma\overset{*}{\to}w$ be a derivation in\emph{ $G$}. In
this derivation we can replace all the uses of the production $A_{i}\to\zeta$
with the sequence $A_{i}\to X_{\zeta}\to\zeta$ and the uses of the
production $\zeta\to B_{j}$ with the sequence $\zeta\to Y_{\zeta}\to B_{j}$
and get a derivation that is valid in $G$, since all the other productions
are common between the two grammars. Thus we have proved that for
all \emph{$\gamma\in N^{+}$} we can convert a valid derivation from
$G$, $\gamma\overset{*}{\to}w$ into a valid derivation in $G'$
and in particular this is true also for $S\overset{*}{\to}_{G}w$,
which can be converted into $S\overset{*}{\to}_{G'}w$ and which furthermore
can be converted into $S'\to S\overset{*}{\to}_{G'}w$ and therefore
$L(G)\subseteq L(G')$. 

Conversely, let \emph{$\gamma\in N^{+}$} and $\gamma\overset{*}{\to}w$
be a valid derivation in $G'$ then whenever we use the rule $A_{i}\to X_{\zeta}$
in this derivation $\gamma\overset{*}{\to}\alpha A_{i}\beta\to\alpha X_{\zeta}\beta\overset{*}{\to}w$
let $\gamma\overset{*}{\to}\alpha A_{i}\beta\to\alpha X_{\zeta}\beta\overset{*}{\to}\delta X_{\zeta}\eta\to\delta\zeta\eta\overset{*}{\to}w$
be the place where the $X_{\zeta}$ that occurs between $\alpha$
and $\beta$ is rewritten (used in the lhs of a production, even if
it rewrites to the same symbol) for the first time. Then necessarily
the $X_{\zeta}\to\zeta$ rule is the one that applies since it is
the only one that has $X_{\zeta}$ in the lhs. Furthermore, as consequence
of Lemma 1 we have that $\alpha\overset{*}{\to}\delta$ and $\beta\overset{*}{\to}\eta$
are valid derivations in $G'$. Therefore we can bring the production
$X_{\zeta}\to\zeta$ right before the use of $A_{i}\to X_{\zeta}$
as follows $\gamma\overset{*}{\to}\alpha A_{i}\beta\to\alpha X_{\zeta}\beta\to\alpha\zeta\beta\overset{*}{\to}\delta\zeta\beta\overset{*}{\to}\delta\zeta\eta\overset{*}{\to}w$
and still have a valid derivation in $G'$. We can repeat this procedure
for all places where rules of the form $A_{i}\to X_{\zeta}$ appear.
Similarly for the uses of the productions of the type $\zeta\to Y_{\zeta}$
in a derivation $\gamma\overset{*}{\to}w$ we can bring the production
that rewrites $X_{\zeta}$ ($Y_{\zeta}\to B_{j}$) right after as
follows: change $\gamma\overset{*}{\to}\alpha\zeta\beta\to\alpha Y_{\zeta}\beta\overset{*}{\to}\delta Y_{\zeta}\eta\to\delta B_{j}\eta\overset{*}{\to}w$
into $\gamma\overset{*}{\to}\alpha\zeta\beta\to\alpha Y_{\zeta}\beta\to\alpha B_{j}\beta\overset{*}{\to}\delta B_{j}\beta\overset{*}{\to}\delta B_{j}\eta\overset{*}{\to}w$,
because from Lemma 1 we have that $\alpha\overset{*}{\to}\delta$
and $\beta\overset{*}{\to}\eta$. We can repeat this procedure for
all places where rules of the form $\zeta\to X_{\zeta}$ appear. In
the new modified derivation we can replace all the uses of the sequence
$A_{i}\to X_{\zeta}\to\zeta$ with the production $A_{i}\to\zeta$
and the sequence $\zeta\to Y_{\zeta}\to B_{j}$ with the production
$\zeta\to B_{j}$ and obtain a derivation that is valid in $G$ since
all the other productions are common between the two grammars. Thus,
we have proved that for all \emph{$\gamma\in N^{+}$} we can convert
a valid derivation from $G'$ ,$\gamma\overset{*}{\to}w$ into a valid
derivation in $G$. For a derivation and $S'\overset{*}{\to_{G'}}w$
we have that necessarily $S'\to S\overset{*}{\to_{G'}}w$ since $S'\to S$
is the only production that rewrites $S'$ but since $S\in N^{+}$,
it follows that we can use the previous procedure in order to convert
$S\overset{*}{\to_{G'}}w$ into a derivation $S\overset{*}{\to_{G}}w$
which proves that $L(G')\subseteq L(G)$. Hence, we have proved that
the grammars are equivalent, i.e., $L(G)=L(G')$.

It is obvious that all the productions of the form $A\to\zeta$ and
$\zeta\to B$ , $\zeta\notin N$ respect \emph{strong-uniqueness}.
Furthermore if the initial grammar has some restrictions on the composition
of the $\zeta\notin N$ that appear in the productions of type 2 and
3 they are still maintained since $\zeta$ is left unchanged in the
productions of the new grammar and the only other types of productions
introduced are Renamings which do not belong to either type 2 or 3. 

$\Box$

\subsection*{Minimality}

We can ask the question whether we can find even simpler types of
structural elements that can generate the full power of Turing Machines.
Two of our operators, RSS and SS require that the size of the production
(\emph{$|lhs|+|rhs|$}) is 3. We can ask whether we can do it with
size 2. This is answered negatively by the following proposition.

\textbf{Proposition (Minimality) }\emph{If we impose the restriction
that $|lhs|+|rhs|\leq2$ for all the productions then we can only
generate languages that are finite sets Terminals, with the possible
addition of the empty string $\epsilon$.}

\emph{Proof . }The only possible productions under this constraint
are:
\begin{enumerate}
\item $A\to a$
\item $A\to\epsilon$
\item $A\to B$
\item $AB\to\epsilon$
\item $Aa\to\epsilon$
\end{enumerate}
All the derivations that start from $S$, either keep on rewriting
the current NonTerminal, because there is no production which increases
the size, or rewrite the current NonTerminal into a Terminal or $\epsilon$.

$\Box$

Another possible question to ask is whether we can find a smaller
set of structural elements set or at least equally minimal. The following
theorem is a known result by Savitch \cite{Savitch 1973}.

\textbf{Theorem 7 (Strong-CFG-ASNF-Savitch). }\emph{(Savitch 1973
\cite{Savitch 1973}) Let $G=(N,T,S,R)$, }$\epsilon\notin L(G)$\emph{
be a General Grammar then there exists a grammar $G'$ such that $L(G)=L(G')$
and $G'$ contains only rules of the following type:}
\begin{enumerate}
\item \emph{$AB\to\epsilon$}
\item \emph{$A\to BC$}
\item \emph{$A\to a$}
\end{enumerate}
The Savitch theorem has three types of rules TERMINAL (type 3), SS
(type 2) and ANNIHILATE2 (type 1). However no \emph{strong-uniqueness}
has been enforced; It it were it to be enforced, then one more type
of rule, REN will be needed. Furthermore if we would not insist on
uniqueness all the Renamings in the ASNF could be eliminated too (Renamings
elimination is a well known technique in the theory of formal languages
\cite{Salomaa 1985}) . However this will make it very difficult to
make explicit the Abstractions. Therefore it can be argued that ASNF
and the Savitch Normal Form have the same number of fundamental structural
operations with the crucial difference between the two being the replacement
of the RSS with ANNIHILATE2. The Reverse SuperStructuring seems a
more intuitive structural operation however, at least from the point
of view of this paper.. Next we present for completeness the version
of Savitch theorem where \emph{strong-uniqueness} is enforced for
rules of type 2 and 3. 

\textbf{Theorem 8 (Strong-GEN-ASNF-Savitch).}\emph{ Let $G=(N,T,S,R)$,
}$\epsilon\notin L(G)$\emph{ be a General Grammar then there exists
a grammar $G'$ such that $L(G)=L(G')$ and $G'$ contains only rules
of the following type:}
\begin{enumerate}
\item \emph{$A\to B$}
\item \emph{$A\to BC$ - and this is the only rule that has $BC$ in the
rhs and this is the only rule that has $A$ in the lhs. (strong-uniqueness)}
\item \emph{$A\to a$ - and this is the only rule that has $a$ in the rhs
and this is the only rule that has $A$ in the lhs. (strong-uniqueness)}
\item \emph{$AB\to\epsilon$ - and this is the only rule that has $AB$
on the lhs. (uniqueness) }
\end{enumerate}
\emph{Proof. }By the original Savitch theorem \cite{Savitch 1973}
any grammar can we written such that it only has rules of the type
\begin{enumerate}
\item \emph{$AB\to\epsilon$}
\item \emph{$A\to BC$}
\item \emph{$A\to a$}
\end{enumerate}
The rules of the form $AB\to\epsilon$ observe uniqueness by default
since the only rules that have more than one symbol in the lhs are
of the form $AB\to\epsilon$.

Then we can use the constructions in Lemma 2 on this grammar in order
to enforce \emph{strong-uniqueness} for SuperStructuring (SS) \emph{$A\to BC$}
and Terminals (TERMINAL) $A\to a$ at the potential expense of introducing
Renamings (REN).

$\Box$

\subsection*{Grow \& Shrink theorems}

\textbf{Theorem 9. }\emph{Let $G=(N,T,S,R)$, }$\epsilon\notin L(G)$\emph{
be a General Grammar in the Strong-GEN-ASNF-Savitch, i.e., all the
productions are of the following form: }
\begin{enumerate}
\item \emph{$A\to B$}
\item \emph{$A\to BC$ - and this is the only rule that has $BC$ in the
rhs and this is the only rule that has $A$ in the lhs. (strong-uniqueness)}
\item \emph{$A\to a$ - and this is the only rule that has $A$ on the lhs
and there is no other rule that has $a$ on the rhs. (strong uniqueness) }
\item \emph{$AB\to\epsilon$ - and this is the only rule that has $AB$
on the lhs. (uniqueness) }
\end{enumerate}
\emph{Then for any derivation $w$ such that $\gamma\overset{*}{\to}w$
, in $G$, $\gamma\in N^{+}$ there exists a derivation $\gamma\overset{*}{\to}\mu\overset{*}{\to}\nu\overset{*}{\to}w$
such that $\mu\in N^{+}$, $\nu\in N^{*}$ and $\gamma\overset{*}{\to}\mu$
contains only rules of type 1 and 2 (REN, SS), $\mu\overset{*}{\to}\alpha$
contains only rules of the type 4 (ANNIHILATE2) and $\nu\overset{*}{\to}w$
contains only rules of type 3 (TERMINAL).}

\emph{Proof. }Based on Lemma 3 (presented next) we can change the
derivation \emph{$\gamma\overset{*}{\to}w$ }into a derivation \emph{$\gamma\overset{*}{\to}\nu\overset{*}{\to}w$
}such that the segment \emph{$\nu\overset{*}{\to}w$ }contains only
rules of type 3 (TERMINAL) and \emph{$\gamma\overset{*}{\to}\nu$
}contains only rules of type 1, 2 or 4 (REN, SS, ANNIHILATE2) . Therefore
the only thing we still have to prove is that we can rearrange \emph{$\gamma\overset{*}{\to}\nu$}
into \emph{$\gamma\overset{*}{\to}\mu\overset{*}{\to}\nu$ }such that
$\gamma\overset{*}{\to}\mu$ contains only rules of type 1 or 2 (REN,
SS) and $\mu\overset{*}{\to}\nu$ contains only rules of the type
4 (ANNIHILATE2). 

Let \emph{$\gamma\in N^{+}$}, and $w$ such that\emph{ $\gamma\overset{*}{\to}\nu\overset{*}{\to}w$}
is a derivation in\emph{ $G$} , the segment \emph{$\nu\overset{*}{\to}w$
}contains only rules of type 3 (TERMINAL) and \emph{$\gamma\overset{*}{\to}\alpha$
}contains only rules of type 1, 2 or 4 (REN, SS, ANNIHILATE2). If
the derivation already satisfies the condition of the lemma then we
are done. Otherwise examine in order the productions in $\gamma\overset{*}{\to}\nu$
from end to the beginning until we encounter the first rule of type
4: $AB\to\epsilon$ that violates the condition required by the theorem
and at least one production of the type 1 or 2 has been used after
it (otherwise we have no violation). More exactly, prior to it only
rules of type 1 or 2 were used and at least one such rule was used.
That is $\gamma\overset{*}{\to}\alpha AB\beta\to\alpha\beta\overset{+}{\to}\mu'\overset{*}{\to}\nu$
and only rules of the type 1 or 2 have been used in $\alpha\beta\overset{+}{\to}\mu$,
and only rules of the type 4 have been used in $\mu'\overset{*}{\to}\nu$.
Because only rules of the type 1 or 2 (which never have more than
one symbol in the lhs) have been used in $\alpha\beta\overset{+}{\to}\mu$
it follows that there exists $\mu_{1},\mu_{2}\in N^{*}$such that
$\alpha\overset{*}{\to}\mu_{1}$ and $\beta\overset{*}{\to}\mu_{2}$
and $\mu=\mu_{1}\mu_{2}$. Therefore we can rearrange the rewriting
$\gamma\overset{*}{\to}\alpha AB\beta\to\alpha\beta\overset{+}{\to}\mu$
into $\gamma\overset{*}{\to}\alpha AB\beta\overset{*}{\to}\mu_{1}AB\beta\overset{*}{\to}\mu_{1}AB\mu_{2}\to\mu_{1}\mu_{2}=\mu$
. In this way we have obtained a derivation for $\mu$ in $G$ that
violates the conclusion of the lemma in one place less than the initial
derivation. Since there is a finite number of steps in a derivation
and therefore a finite number of places where the constraints can
be violated it can be inferred that after a finite number of applications
of the above-described {}``swapping'' procedure we will obtain a
derivation which satisfies the rules of the theorem.

$\Box$

\textbf{Lemma 3.} \emph{Let $G=(N,T,S,R)$ be a general (unrestricted)
grammar that contains only rules of the form:}
\begin{enumerate}
\item $\alpha\to\beta$, $\alpha\in N^{+}$,$\beta\in N^{*}$
\item $A\to a$ 
\end{enumerate}
\emph{Then for any derivation $w$ such that $\gamma\overset{*}{\to}w$
, in $G$, $\gamma\in N^{+}$ there exists a derivation $\gamma\overset{*}{\to}\nu\overset{*}{\to}w$
such that $\nu\in N^{+}$ and $\gamma\overset{*}{\to}\nu$ contains
only rules of type 1 and $\nu\overset{*}{\to}w$ contains only rules
of type 2 (TERMINAL).}

\emph{Proof. }Let $w$ such that $\gamma\overset{*}{\to}w$ \emph{in
$G$, $\gamma\in N^{+}$}. If the derivation already satisfies the
condition of the lemma then we are done. Otherwise examine in order
the productions $\gamma\overset{*}{\to}w$ until we encounter a rule
of type , say it is $A\to a$ , such that there are still rules of
type 1 used after it $\gamma\overset{*}{\to}\alpha A\beta\to\alpha a\beta\overset{*}{\to}w$.
Because none of the rules in the grammar contain terminals in their
lhs it follows that there exists $w_{1},w_{2}\in T^{*}$ such that
$\alpha\overset{*}{\to}w_{1}$ and $\beta\overset{*}{\to}w_{2}$ and
$w=w_{1}aw_{2}$. Therefore we can rearrange the rewriting $\gamma\overset{*}{\to}\alpha A\beta\to\alpha a\beta\overset{*}{\to}w$
into $\gamma\overset{*}{\to}\alpha A\beta\overset{*}{\to}w_{1}A\beta\overset{*}{\to}w_{1}Aw_{2}\to w_{1}aw_{2}=w$.
In this way we have obtained a derivation for $w$ in $G$ that violates
the conclusion of the lemma in one place less than the initial derivation.
Since there is a finite number of steps in a derivation and therefore
a finite number of places where the constraints can be violated it
can be inferred that after finite number of application of the above-described
{}``swapping'' procedure we will obtain a derivation which satisfies
the rules of the lemma.

$\Box$

\textbf{Theorem 6. }\emph{Let $G=(N,T,S,R)$, }$\epsilon\notin L(G)$\emph{
be a General Grammar. Then we can convert such a grammar into the
Strong-GEN-ASNF i.e., such that all the productions are of the following
form: }
\begin{enumerate}
\item \emph{$A\to B$}
\item \emph{$A\to BC$ - and this is the only rule that has $BC$ in the
rhs and this is the only rule that has $A$ in the lhs. (strong-uniqueness)}
\item \emph{$A\to a$ - and this is the only rule that has $A$ on the lhs
and there is no other rule that has $a$ on the rhs. (strong uniqueness) }
\item \emph{$AB\to C$ - and this is the only rule that has $C$ in the
rhs and this is the only rule that has $AB$ in the lhs. (strong-uniqueness)}
\end{enumerate}
\emph{And furthermore for any derivation $w$ such that $\gamma\overset{*}{\to}w$
, in $G$, $\gamma\in N^{+}$ there exists a derivation $\gamma\overset{*}{\to}\mu\overset{*}{\to}\nu\overset{*}{\to}w$
such that $\mu\in N^{+}$, $\nu\in N^{*}$ and $\gamma\overset{*}{\to}\mu$
contains only rules of type 1 and 2 (REN, SS), $\mu\overset{*}{\to}\alpha$
contains only rules of the type 1, more particularly only Reverse
Abstractions and type 4 (REN(RABS), RSS) and $\nu\overset{*}{\to}w$
contains only rules of type 3 (TERMINAL).}

\emph{Proof Sketch. }By Theorem 6 we can convert the grammar $G$
into a grammar $G'$ in Strong-GEN-ASNF-Savitch. Then we can convert
such a grammar into a grammar $G''$ in Strong-GEN-ASNF as follows:
for all $\{A_{i}B_{i}\to\epsilon\}_{i=1,n}$ introduce new NonTerminals
$\{X_{A_{i}B_{i}}\}_{i=1,n}$ and add $A_{i}B_{i}\to X_{A_{i}B_{i}}$
and $X_{A_{i}B_{i}}\to E$ and eliminate the original productions
$\{A_{i}B_{i}\to\epsilon\}_{i=1,n}$. Furthermore, for all NonTerminals
$X\neq E$ in the new grammar, add new NonTerminals $X_{XE}$, $X_{EX}$
and $X_{EE}$ and the production rules $XE\to X_{XE}$, $EX\to X_{EX}$,
$X_{XE}\to X$, $X_{EX}\to X$, $EE\to X_{EE}$ and $X_{EE}\to E$.
We can easily show using techniques already developed that the new
grammar will generate the same language as the previous one and that
it respects the \emph{strong-uniqueness} for rules of type 2, 3 and
4. 

Furthermore if we take a derivation for a string $w$ in the grammar
$G'$ in the Strong-GEN-ASNF-Savitch $S\overset{*}{\to_{G'}}w$ then
from Theorem 7 we know that we can convert it into a derivation that
uses first only REN and SS in the phase 1, then only ANNIHILATE2 in
phase 2 and finally only TERMINAL in phase 3. We can take such a derivation
and replace the usage of the ANNIHILATE2 productions in phase 2 with
productions as above in order to get a derivation in $G''$. Note
that the productions introduced above are meant to transform the ANNIHILATE2
into rules that follow the Strong-GEN-ASNF, and the only types of
rules that we have introduced are RSS with strong-uniqueness holding
and REN of the RABS type. This proves the theorem.

$\Box$

\subsection*{Further Discussions and a Conjecture}

We have therefore proved that for each General Grammar $G$ we can
transform it both in a Strong-GEN-ASNF-Savitch and a Strong-GEN-ASNF
such that the derivation (explanation in Structural Induction terminology)
of any terminal string $w$ can be organized in three phases such
that: In Phase 1 we use only productions that grow (or leave the same)
the size of the intermediate string; In Phase 2 we use only productions
that shrink (or leave the same) the size of the intermediate string
and in Phase 3 we use only TERMINAL productions.

Naively, at first sight, it may seem that this is a way to solve the
halting problem and therefore maybe some mistake has been made in
the argument. However this is not the case, as the key question is
when to stop expanding the current string and start shrinking and
this problem still remains. In a certain sense these two theorems
are a way to give a clear characterization of the issue associated
with solving the halting problem: namely that of knowing when to stop
expanding. In the case of Grammars in arbitrary forms the issue is
a little bit more muddled as we can have a succession of grow and
shrink phases but we have shown that if the form is constrained in
a certain ways then we only need one grow and one shrink. Note also
that during the grow phase in both theorems we are only using context
free productions.

\subsection*{Structural Induction and the fundamental structural elements}

In this section we will review the Structural Induction process in
the light of the concepts and results obtained for Generative Grammars
and discuss the role of each of the operators. Then we move on to
make some more connections with already existing concepts in Statistics
and Philosophy and show how the ASNF affords for very precise characterizations
of these concepts.

In the context of Generative Grammars Structural Induction is concerned
with the following question: Given a sequence of observations $w$
we attempt to find a theory (grammar) that explains $w$ and simultaneously
also the explanation (derivation) $S\overset{*}{\to}w$. In a local
way we may think that whenever we have a production rule $l\to r$
that $l$ explains $r$. In a bottom up - data driven way we may proceed
as follows: First introduce for every Observable $a$ a production
$A\to a$. These are just convenience productions that bring the observables
into the realm of internal variables in order to make everything more
uniform. The association is unique (one to one and onto) and once
we have done it we can forget about the existence of Observables (Terminals).
This is only the role of the the TERMINAL productions, and for this
reason we will not mention them in future discussions as they are
not true structural operations. With this in mind if we are to construct
a theory in the GEN-ASNF we can postulate any of the following laws:
\begin{enumerate}
\item SS - $A\to BC$ - SuperStructuring. Takes two internal variables $B$
and $C$ that occur within proximity of each other (adjacent) and
labels the compound. From now on the shorter name $A$ can be used
instead on the compound. This is the sole role of SuperStructuring
- to give a name to a bigger compound in order to facilitate shorter
explanations at latter stages.
\item ABS - $A\to B|C$ - Abstraction. Makes a name for the occurrence of
either of the variables $B$or $C$. This may allow for the potential
bundling of two productions, the first one involving $B$ and the
other one involving $C$ while otherwise being the same, into a production
involving only $A$. The role of Abstraction is to give a name to
a group of entities (we have chosen two for simplicity) in order to
facilitate more general explanations at latter stages which in turn
will produce more compact theories.
\item RSS - $AB\to C$ - Reverse SuperStructuring - invent up to two new
internal variables (may also use already existing ones) which if they
occur within proximity of each other together they {}``explain''
the internal variable $C$.
\item RABS - $A\to C$, $B\to C$ - Reverse Abstraction - invent new internal
variables (may also use already existing ones) such that either of
them can {}``explain'' the internal variable $C$ (we have chosen
two variables for simplicity). 
\end{enumerate}
In the next subsection we review two possible reasons for creating
hidden variables and we identify them with Reverse Abstraction and
Reverse SuperStructuring. Because in our GEN-ASNF we do not have other
types of reasons for creating Hidden Variables as all the other production
introduce convenience renamings only we can infer under the Computationalistic
Assumption that these two are the essential reasons for postulating
Hidden Variables and any other reasons must be derivative. This produces
a definite structural characterization of the rationales behind inventing
Hidden Variables.

\subsubsection*{Reasons for postulating Hidden Variables}

There are at least two types of reasons for creating Hidden Variables:
\begin{enumerate}
\item (\textbf{OR} type) - {[}multiple alternative hidden causes{]} The
OR type corresponds to the case when some visible effect can have
multiple hidden causes $H1\to Effect$, $H2\to Effect$ . This case
corresponds in our formalism to the notion of Reverse Abstraction.
One typical example of this is: The grass is wet, hence either it
rained last night or the sprinkler was on. In the Statistical literature
the models that use these types of hidden variables are known as mixture
models \cite{Lauritzen 1996 5}.
\item (\textbf{T-AND} type) - {[}multiple concurrent hidden causes{]} The
T-AND type, i.e., topological AND type, of which the AND is a particular
case. This corresponds to the case when one visible effect has two
hidden causes that both have to occur within proximity (hence the
Topological prefix) of each other in order for the visible effect
to be produced. $H1H2\to Effect$. This corresponds in our formalism
to the case of Reverse SuperStructuring. One example of this case
is the Annihilation of an electron and a positron into a photon. $e^{+}e^{-}\to\gamma$
as illustrated by the following Feynman diagram. In the diagram the
Annihilation is followed by a Disintegration which in our formalism
will be represented by a SuperStructure $\gamma\to e^{+}e^{-}$. Note
that the electron positron annihilation is a RSS and not an ANNIHILATE2
as in the Savitch type of theorems. In a certain sense one of the
arguments for using RSS versus ANNIHILATE2 is also the fact that physical
processes always have a byproduct despite carrying potentially misleading
names such as annihilation, or the byproduct being energetic rather
than material. Nevertheless, since our reasons for preferring GEN-ASNF
versus GEN-ASNF-Savitch alternative are at best aesthetic / intuitive
reasons it should always be kept in mind as a possible alternative.
In the Statistical / Graphical Models literature the particular case
of AND hidden explanations is the one that introduces edges between
hidden variables in the depedence graph \cite{Elidan and Friedman 2005},
\cite{Lauritzen 1996 5}, \cite{Pearl 1988 5}.
\end{enumerate}
\begin{figure}
\begin{centering}
\includegraphics{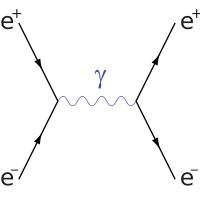}
\par\end{centering}

\caption{Feynman diagram for the electron positron Annihilation into a photon
followed by photon Disintegration into an electron positron pair again}
\end{figure}

Our analysis of the Turing equivalent formalism of Generative Grammars
written in the ASNF has evidenced them as the only needed types and
we can infer under the Computationalistic Assumption that these are
the only two essential reasons for postulating Hidden Variables and
any other reasons must be derivative.

\subsubsection*{Radical Positivism}

Since RSS and RABS involve the postulation of hidden variables, and
we have discussed the perils associated with it, one alternative is
to choose to use only Abstraction and SuperStructuring. We propose
that this in fact this characterizes the radical positivist \cite{Ayer 1936}
stance which allows for empirical laws only. After we rule out RSS
and RABS since they may introduce Hidden Variables we are left with
ABS and SS and this produces our proposed characterization of the
radical positivist position and what we mean by empirical laws under
the Computationalistic Assumption. The internal variables created
by Abstraction and SuperStructuring are going to be just convenient
notations for aggregates of input data but nothing else: SuperStructuring
is just convenient labeling of already existing structures for the
sake of brevity and Abstraction on the other hand aggregates a group
of variables into a more general type so that we can produce more
encompassing laws but with coarser granularity. An explanation of
a stream of observations $w$ in the Radical Positivist theory of
the world (or least a piece of it) will look like the picture in Figure
\ref{fig:Theory-of-the} - top. In this picture the atoms are the
observations and the theory of the universe is mainly a theory of
how the observables are grouped into classes (Abstractions) and how
smaller chunks of observations are tied together into bigger ones
(SuperStructures). The laws of the radical positivist theory are truly
empirical laws as they only address relations among observations. 

However using only these two operators we cannot attain the power
of Turing Machines. More precisely, the resulting types of grammars
will be a strict subset of Context Free Grammars, (since CFG contain
SS, and REN(ABS+RABS)). Next we will examine how any theory of the
world may look like from the most general perspective when we do allow
Hidden Variables.

\subsubsection*{General theories of the world}

\begin{figure}
\begin{centering}
\includegraphics[scale=0.5]{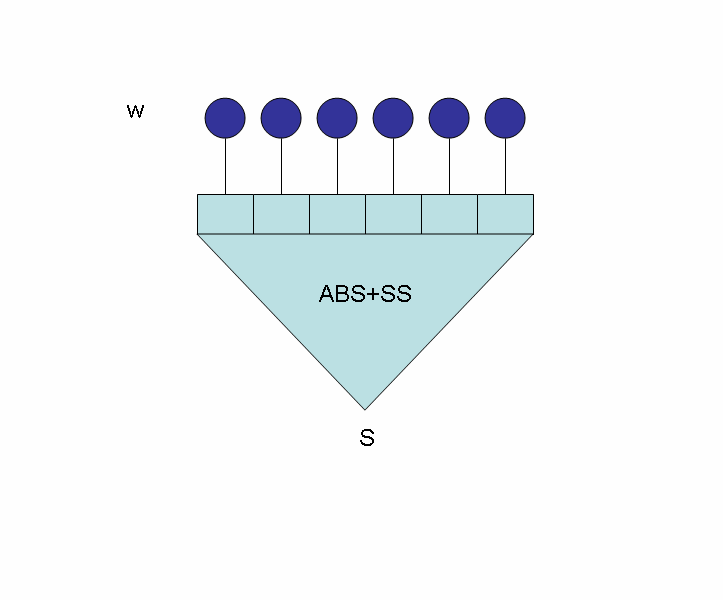}
\par\end{centering}

\begin{centering}
\includegraphics[scale=0.5]{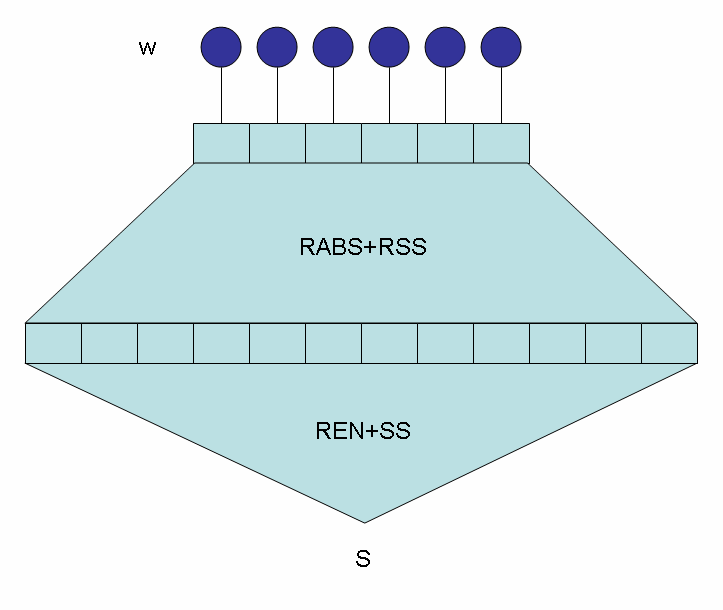}
\par\end{centering}

\caption{\label{fig:Theory-of-the}Theory of the world: (top) - Radical Positivist,
(bottom) - with Hidden Variables}
\end{figure}

An explanation of a stream of observations $S\overset{*}{\to}w$ in
the more general hidden variable theory of the world is illustrated
in Figure \ref{fig:Theory-of-the} - bottom. The atoms are the hidden
variables and their organization is again in turn addressed by the
Abstraction and SuperStructuring but also Reverse Abstraction. This
part is basically a context free part since all the productions are
context free. Observations are a derivative byproduct obtained from
a richer hidden variable state description by a reduction: either
of size - performed by Reverse SuperStructuring or of information
- performed by Reverse Abstraction.

The hidden variables theory of the world picture is an oversimplification
of the true story, in general we may have a set of alternations of
REN+SS and RABS+RSS rather than just one. However as we have shown
in Theorem 8 we can turn any grammar into a grammar in Strong-GEN-ASNF
such that any explanation done in three phases only (as illustrated
in Figure \ref{fig:Theory-of-the}): 
\begin{enumerate}
\item a growth phase where we use only REN+SS 
\item a shrink phase where we use only RABS+RSS 
\item a phase where we only rewrite into Terminals. 
\end{enumerate}
Whether additional separations can be made, e.g., if we can push all
the RABS from the first phase into the second phase and make the first
phase look like the one in the radical positivist story (i.e., using
only ABS+SS) is a topic of further investigation. We take this opportunity
to proposed it as a conjecture.

\textbf{Conjecture. }\emph{Let $G=(N,T,S,R)$, }$\epsilon\notin L(G)$\emph{
be a General Grammar. Then we can convert such a grammar into the
Strong-GEN-ASNF i.e., such that all the productions are of the following
form: }
\begin{enumerate}
\item \emph{$A\to B$}
\item \emph{$A\to BC$ - and this is the only rule that has $BC$ in the
rhs and this is the only rule that has $A$ in the lhs (strong-uniqueness).}
\item \emph{$A\to a$ - and this is the only rule that has $A$ on the lhs
and there is no other rule that has $a$ on the rhs. (strong uniqueness) }
\item \emph{$AB\to C$ - and this is the only rule that has $C$ in the
rhs and this is the only rule that has $AB$ in the lhs (strong-uniqueness).}
\end{enumerate}
\emph{And furthermore for any derivation $w$ such that $\gamma\overset{*}{\to}w$
, in $G$, $\gamma\in N^{+}$ there exists a derivation $\gamma\overset{*}{\to}\mu\overset{*}{\to}\nu\overset{*}{\to}w$
such that $\mu\in N^{+}$,$\nu\in N^{*}$ and $\gamma\overset{*}{\to}\mu$
contains only rules of type 1, more particularly only Abstractions,
and type 2 (ABS,SS), $\mu\overset{*}{\to}\alpha$ contains only rules
of the type 1, more particularly only Reverse Abstractions, and type
4 (RABS,RSS) and $\nu\overset{*}{\to}w$ contains only rules of type
3 (TERMINAL).}

In a certain sense the conjecture can be seen as way to try to salvage
as much as we can from the Radical Positivist position by saying that
in principle it is right with the caveat that the atoms should be
internal(hidden) variables rather than observations. If we were God
(here used in the scientifico-philosophical sense - i.e., somebody
which knows the laws of the universe and has access to the full hidden
state of it) then we would be able to stop our explanation of the
current state after phase 1, which would use only Abstractions and
SuperStructures (the Radical Positivist position). However since we
are mere mortals and all we are able to perceive are Observables which
are just a simplified small reflection of the current hidden state
of the world there is a the need for reduction operations: reduction
in size - performed by Reverse SuperStructuring and reduction of information
- performed by Reverse Abstraction. This is then followed by the one
to one correspondence between some internal variable and Observables
(Terminals in grammar terminlogy).

Our main claim so far is that we can rewrite any General Grammar in
GEN-ASNF, that is using only ABS, SS, RABS and RSS. In the next section
we will examine a statement that was made by the philosopher David
Hume more that 200 years ago in the light of the GEN-ASNF theorem
and propose that they are more or less equivalent under the Computationalistic
Assumption. We then examine the developments that occured in the meantime
in order to facilitate our proof of Hume's claim.

\subsubsection*{Hume principles of connexion among ideas}

\emph{{}``I do not find that any philosopher has attempted to enumerate
or class all the principles of association {[}of ideas{]}. ... To
me, there appear to be only three principles of connexion among ideas,
namely, }\textbf{\emph{Resemblance}}\emph{, }\textbf{\emph{Contiguity}}\emph{
in time or place, and }\textbf{\emph{Cause }}\emph{and }\textbf{\emph{Effect''}}\emph{
- David Hume, Enquiry concerning Human Understanding, III(19), 1748.
\cite{Hume 1993}}

If we are to substitute Resemblance for Abstraction (as resemblance/similarity
is the main criterion for abstraction), Contiguity in time or place
for SuperStructuring (as the requirement for SuperStructuring is proximity
- in particular spatial or temporal) and Cause and Effect for the
two types of hidden variable explanations then the GEN-ASNF theorem
is the proof of a more the two hundred years old claim. The main developments
that have facilitated this result are: 
\begin{enumerate}
\item The Church-Turing thesis -1936 \cite{Turing 1936} (i.e., the Computationalistic
Assumption) which allowed us to characterize what a theory of the
world is, i.e., an entity expressed in a Turing equivalent formalism.
\item The Generative Grammars - 1957 \cite{Chomsky 1957} the development
of the Compositional formalism of Generative Grammars which is Turing
equivalent.
\item Developments in understanding the structure of Generative Grammars
- The Kuroda Normal Form 1964 \cite{Kuroda 1964} and General Kuroda
Normal Form \cite{Salomaa 1985}.
\item The GEN-ASNF theorems proved in this paper. 
\end{enumerate}
Furthermore the elucidation of the two types of causal explanation
(alternative - Reverse Abstraction (RABS) and topological conjunctive
- Reverse SuperStructuring (RSS)) is an additional merit of GEN-ASNF
theorem. It should be mentioned also that we became aware of Hume's
claim after we have already stated the GEN-ASNF theorem but prior
to it's proof in the full generality. We were led to it in a certain
sense by similar intuitions and investigations into the nature of
the Structural Induction process as David Hume's. Once aware of it,
this became an additional supporting argument for the fact that the
proof may be a feasible enterprise. 
\end{document}